\def\year{2019}\relax
\documentclass[letterpaper]{article} 
\usepackage{aaai19}  
\usepackage{times}  
\usepackage{helvet}  
\usepackage{courier}  
\usepackage{url}  
\usepackage{graphicx}  
\usepackage{wrapfig}  
\usepackage{multicol}
\usepackage[linesnumbered,ruled,vlined]{algorithm2e}
\newcommand{\eat}[1]{}

\begin{document}
%
\title{Automated Test Generation to Detect \\Individual Discrimination in AI Models}

\author{Aniya Agarwal, Pranay Lohia, Seema Nagar, Kuntal Dey, Diptikalyan Saha\\
IBM Research AI\\
India\\
Email: \{aniyaagg,plohia07,senagar3,kuntadey,diptsaha\}@in.ibm.com\\
}
\maketitle
\begin{abstract}
Dependability on AI models is of utmost importance to ensure full
acceptance of the AI systems. One of the key aspects of the dependable
AI system is to ensure that all its decisions are fair and not biased
towards any individual. In this paper, we address the problem of
detecting whether a model has an individual discrimination. Such a
discrimination exists when two individuals who differ only in the
values of their protected attributes (such as, gender/race) while the
values of their non-protected ones are exactly the same, get different
decisions. Measuring individual discrimination requires an exhaustive
testing, which is infeasible for a non-trivial system. In this paper,
we present an automated technique to generate test inputs, which is
geared towards finding individual discrimination. Our technique
combines the well-known technique called symbolic execution along with
the local explainability for generation of effective test cases. Our
experimental results clearly demonstrate that our technique produces
3.72 times more successful test cases than the existing
state-of-the-art across all our chosen benchmarks.

\eat{
In the last few years, there has been a tremendous surge of deep
neural network-based decision systems. As these systems are used in
practical decision making, automated test case needs to be generated
to test such a system. Symbolic or concolic evaluation based
algorithms are successful in systematically generating test inputs to
cover the decision space (e.g. path coverage or branch coverage) of
the traditional programs. While there exists few works that try to
perform symbolic or concolic execution based algorithms for deep
neural networks. These techniques are white box, and are not scalable
as they try to cover structural entities of the neural network which
are really large in numbers. Also, the constraint solvers are not able
to handle such large formula.

In this paper, we use a novel technique for symbolic execution for
uninterpretable models. We generate local explanations using an
off-the-shelf local explainer and use it to traverse the decision
space of the system using dyanmic symbolic evaluation technique. Our
technique is simple, easily implementable and uses existing constraint
solvers. We experimentally demonstrate the effectiveness of our
technique in three modalities - structural data for solving the
individual discrimination problem, and genrating adversarial test
inputs for test and image classification problems.}
 
\end{abstract}

\section{Introduction}
\label{sec:intro}

\emph{Model Bias.} This decade marks the resurgence of Artificial
Intelligence where AI Models have started taking crucial decisions in
a lot of systems - from hiring decisions, approving loans, etc. to
design driver-less cars. Therefore, dependability on AI models is of
utmost importance to ensure wide acceptance of the AI systems. One of
the important aspects of the dependable AI system is whether decisions
are fair and not biased. Bias may be inherent in a decision-making
system in multiple ways. It can exist in the form of group
discrimination~\cite{DISP} where two different groups (e.g., based on
gender/race) gets a varied decision or an individual
discrimination~\cite{THEMIS}.

\emph{Individual discrimination.} In this paper, we address the
problem of detecting whether a model discriminates between two individuals having the values of all their attributes other than the protected ones exactly the same and if such a model yields different decisions for such two individuals.
Such cases of bias have been previously noticed in models such as~\cite{THEMIS} and
caused derogatory consequences to the model generator. Therefore,
detection of such cases becomes crucial and is of utmost importance. Even though the
training data may not contain two instances where such discrimination
is noticed, the model can still show such an unintended behavior. The challenge is, therefore, to evaluate and find that for which all values of non-protected and
protected attributes, the model demonstrates an individual
discrimination behavior.

\emph{Existing Techniques and their drawbacks.} Measuring individual
discrimination requires exhaustive testing, which is infeasible for a
non-trivial system.  The existing technique, such as THEMIS~\cite{THEMIS}
generates a test suite to determine if and how much individual
discrimination is present in the model. Their approach selects
random values from the domain for all attributes to determine if the
system discriminates amongst the individuals.  Even though such techniques are
applicable for any black-box system, our experiments demonstrate that
they miss many such combinations of non-protected attribute values for which
the individual discrimination may exist. Some of the random
inputs may follow the same execution path in the system having the same effect
on the output.

\emph{Our approach.} There exists symbolic
evaluation~\cite{DART,CUTE,EXE} based techniques to automatically
generate test inputs by systematically exploring
different execution paths in the program. Such methods avoid generation of such
inputs which tend to explore the same paths. Such techniques are
essentially white-box and leverage the capabilities of constraint solvers~\cite{Z3} to create test inputs automatically. Symbolic execution starts with a random input and
analyzes the path to generate a set of path constraints (i.e. conditions on the
input attributes) and iteratively toggles (or negates) the constraints in
the path to generate a set of new path constraints. It then solves the resultant
path constraints using a constraint solver to generate a new input which
can possibly take the control to the new path as explained using an example in Section~\ref{sec:back}.  Our idea is to
use such a dynamic symbolic evaluation to generate test inputs which
can potentially lead to uncovering individual discrimination. However, existing such techniques have been used to generate inputs for procedural programs which are interpretable. Our main challenge is to apply such technique for un-interpretable models.  Note that, similar
to THEMIS, our goal is to build a black box and scalable solution, which can be applied efficiently on varied models.

\emph{Challenges.} There exists a few works which try to use symbolic
evaluation-based techniques for un-interpretable models such as deep
neural networks, although they do not address the problem of finding
individual discrimination in the model. Such techniques are
essentially white-box and try to approximate the functions
(ReLu/Sigmoid) that exist in the network. Therefore, they are catered
towards a specific kind of networks and are not generalizable. Other
test-case generation techniques~\cite{TESTINGDNN,DEEPCONCOLIC} use
coverage criteria (like neuron coverage, sign-coverage, etc.) which
are structure dependent and therefore such techniques suffer from
scalability.

\emph{Solution overview.} In this paper, our key idea is to use the
local explanation as the path in the symbolic execution.  The local
explainer can produce the decision tree corresponding to one input.
The decisions in the decision tree are toggled to generate new
constraints. Below we list several advantages/salient features of our
approach.

\begin{itemize}
\item{Black Box.} Unlike other
  techniques~\cite{DEEPCHECK,DEEPCONCOLIC}, our method is black box as
  the local explainer like LIME~\cite{LIME} handles black-box models.
  This enables us to operate on various types of models including deep neural networks.
\item{Constraints.} It is possible to use an off-the-shelf local
  explainer to generate a linear approximation to the path. The linear
  constraints obtained from the local explainer can be used for the
  symbolic evaluation which won't require any specialized constraint
  solver such as in\cite{TESTINGDNN}.
\item{Data-driven.} We use training data as a seed to start the search. 
\item{Directed and Undirected Search.} Once an individual
  discrimination is found, We perform directed search to uncover many input combinations which can uncover more discrimination. Otherwise,
  we perform an undirected search using symbolic execution to cover paths in the model. 
\item{Optimizations.} The local explainer preserves the important
  constraints for the decision and omits the unnecessary ones. This removes the need for unnecessary toggling of constraints. Our
  algorithm performs the selection of constraints for toggling based
  on its confidence.

\item{Scalability.} Our algorithm systematically traverses paths in
  the feature space by toggling feature related constraints. This makes it scalable, unlike other techniques~\cite{TESTINGDNN} which consider structure-based coverage criteria. 
\end{itemize} 

\emph{Contributions.} Our contributions are listed below:

\begin{itemize}
\item We present a novel technique for finding individual
  discrimination in the model. 
\item We developed a novel combination of dynamic symbolic execution
  and local explanation for generating test cases of uninterpretable models. 
\item We demonstrate the effectiveness of our techniques for several open
  source classification models containing known biases. We compare our technique with the previous algorithm (THEMIS) and demonstrate that we perform better than their approach. 
\end{itemize}

\emph{Outline.} Section~\ref{sec:back} presents the background on
dynamic symbolic execution and local explainability. The following
section presents the algorithm concentrating on various other
challenges in successfully combining the idea of symbolic execution
with the local explanation. Section~\ref{sec:expt} contains the
experimental results. This is followed by the related work in
Section~\ref{sec:related}. Section~\ref{sec:conc} contains the summary,
discussion and future work.

\section{Background}
\label{sec:back}

\subsection{Dynamic Symbolic Execution} Dynamic symbolic execution
(DSE)~\cite{DART,CUTE,EXE} for automated test generation consists of
instrumenting and running a program while collecting path constraint
on inputs from predicates encountered in branch instructions, and of
deriving new inputs from a previous path constraint by a constraint
solver in order to steer next executions toward new program paths.  We
explain the technique using a simple program, shown in
Figure~\ref{fig:ex1}.

\begin{wrapfigure}{l}{2.1cm}
\scriptsize
\begin{verbatim}
f(x, y)
   z = x + y
   if (z > 0)
       p = x -  y
   else
       p = x + 2 * y
   if (p> 0)
       return 1;
   else
       return 0;	
\end{verbatim}
\caption{DSE Example}
\label{fig:ex1}
\end{wrapfigure}

The technique instruments the program such that the instrumented code
performs operation on the symbolic memory which has symbols correspond
to all the variables.  The symbolic memory is initialized with symbols
for the inputs, say $X$ for $x$ and $Y$ and $y$.  The other variable
values are expressed as the expression over input variables. The
symbolic constraints along the path generates the path constraints.
For example, when started with input $(x = 2, y= 3)$, generates the path
constraint $((X+Y>0)\wedge !(X-Y>0))$.  It then selects the last
constraint (depth-first way of path exploration) and creates the path
constraint $((X+Y>0)\wedge (X-Y>0))$ which it solves to get an answer
$X=3, Y=2$. This input, as expected, will go through \texttt{then}
branch of the correspondong toggled condition. In the resultant path,
it will toggle the first branch condition $(X+Y>0)$ and will therefore
generate inputs $(X=-2, Y=1)$ which will take the \texttt{else} branch
of the first condition. This will generate a path constraint
$!(X+Y>0)\wedge !(X + 2 * Y>0)$. It will then solve $!(X+Y>0)\wedge (X
+ 2 * Y>0)$ to generate the fourth input $(X=-2, Y=2)$.

\subsection{Local Explainability}
Local Interpretable Model-agnostic Explanations (LIME) \cite{LIME} consists of explanation techniques that explains the predictions of any classifier or regressor in an interpretable and faithful manner, by approximating it as an interpretable model locally around the prediction. It generates explanation in the form of interpretable models, such as linear models, decision trees, or falling rule lists, which can be easily comprehended by the user with visual or textual artifacts. For generating explanation of an instance, it first converts the representation to an interpretable representation by converting it to a binary vector (0 represents absence of word/image patch). Then it generates data points in the vicinity of the instance by perturbation and learns an interpretable model by minimizing unfaithfulness of the model in approximating the locality of the instance and maximizing local-fidelity and interpretability.

\begin{figure}[t]
\centering
\begin{tabular}{cc}
\includegraphics[height=1.8in]{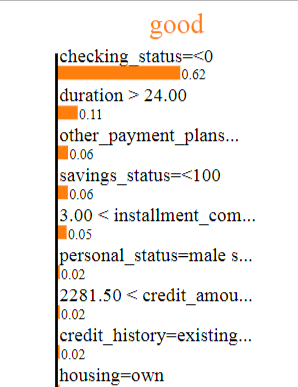} & 
\includegraphics[height=1.8in]{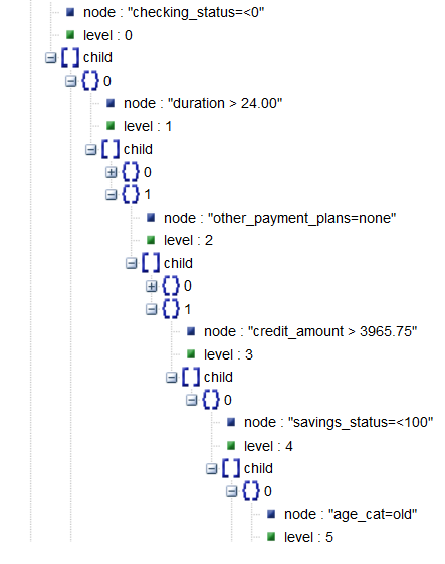}\\
(A) & (B)\\
\end{tabular}
\caption{LIME Explanation (A) and decision tree (B) for an instance on
  German Credit Data}
\label{fig:lime_explanation}
\end{figure}

We use LIME to explain a prediction instance for a model and generate a decision tree as interpretable model by combining the explanations across multiple instances. Figure \ref{fig:lime_explanation} shows an example of LIME explanation for an instance of German Credit Data \cite{german_data}. The data contains bias against young people. It is evident from the figure for explanation that age is an important attribute in the prediction for credit risk of a person.  Figure \ref{fig:lime_explanation} shows a snippet of the decision tree we build using LIME for the German Credit Data. The decision tree is built by combining the explanations for a set of instances from the data.

\section{Algorithm}
\label{sec:algo}

In this section, we present the algorithm for individual
discrimination in an uninterpretable model. We present the algorithm
in two steps.  First, we present a skeleton algorithm of the symbolic
execution which generalizes the symbolic evaluation algorithm from a procedural program to interpretable models. In the next step, we
present the full algorithm of generating test inputs for
uninterpretable models geared towards discovering different inputs for
finding individual discrimination.

\subsection{Generalized Symbolic Execution}  

Algorithm~\ref{algo:gse} presents the algorithm for symbolic execution
such that it can be generalized from programs to models. We explain
the changes in this subsection and in the next subsection explain
explicit changes are required for models.

\begin{algorithm}[htb]
\scriptsize
\SetAlgoLined
\SetAlgoLined\DontPrintSemicolon
count = 0;\\\label{line:l1}
inputs = seed\_test\_inputs()\\\label{line:l2}
priorityQ q = empty; q.addAll(inputs,0); \\\label{line:l3}
\While {count$<$limit \&\& !q.isEmpty()} { \label{line:l4}
  $t$ = deque(q)\\ \label{line:l4}
  check\_for\_error\_condition($t$) \\ \label{line:l4}
  Path p = getPath($t$)\\ \label{line:l5}
  prefix\_pred = true\\ \label{line:l6}
  \ForEach {predicate $c$ in order from top of path}{ \label{line:l8}
  	path\_constraint = prefix\_pred  $\wedge$ toggle (c)\\ \label{line:l9}
	\If {!visited\_path.contains(path\_predicate)}{
	   visited\_path.add(path\_predicate);\\
	   input= solve(path\_constraint)\\ \label{line:l10}
           r = rank(input,S,p)ï \\  \label{line:l10}
           q.add(input,r)\\ \label{line:l11}
	}    
	prefix\_pred = prefix\_pred $\wedge$ c	\label{line:l12}
  }   
  count++ \label{line:l13}
}

\caption{Generalized Dynamic Symbolic Execution}\label{algo:gse}
\end{algorithm}

The first change is related to the inputs to start with. Instead of
starting from a random input, the algorithm finds one or more seed
inputs to start with (Line~\ref{line:l2}). The second change
(Lines~\ref{line:l3},~\ref{line:l10},~\ref{line:l11}) is related to the abstraction of the ranking strategy of selecting which test inputs to
execute next. Note that, now the \texttt{Rank} function will determine
which inputs should be taken next. In the example illustrated in
Section~\ref{sec:back}, we presented a depth-first strategy for
selecting a branch to toggle. And such a decision is taken after each
path is executed. Here, for each path, we consider all the
conditionals to be toggled and associate ranks to them. The third
change is the addition of the check whether the path is already
traversed or not (Line~\ref{line:l8}). Such checks are not required in
symbolic execution for programs as the selection of predicates for
toggling ensures that an already traversed path will not be traversed
again.

\begin{figure}[t]
\centering
\includegraphics[width=5.5cm]{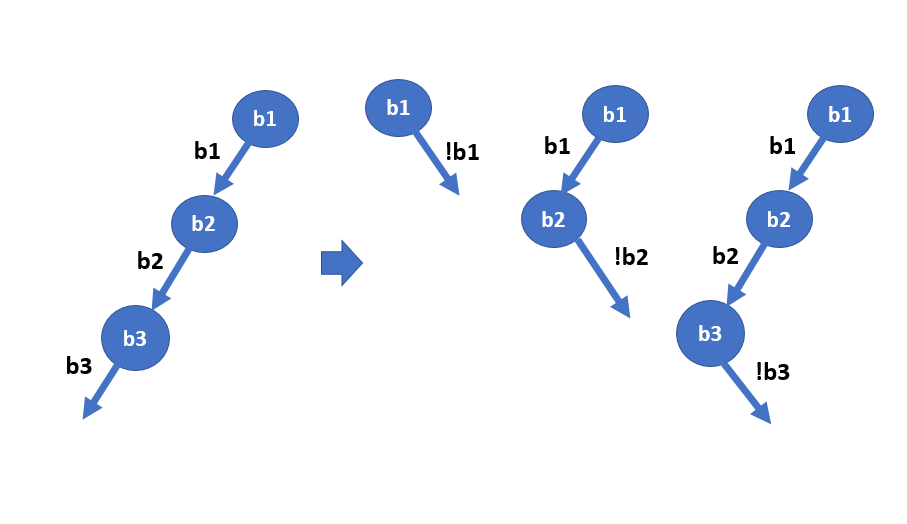} 
\caption{Generation of path constraints}
\label{fig:toggle}
\end{figure}

Note that, the goal of symbolic execution is to explore as much as
paths in the decision space. The generation of constraints
(Lines~\ref{line:l6}-\ref{line:l12} is illustrated in the
Figure~\ref{fig:toggle}. Note that, the other variables, not present
in the constraint, can take any value from the domain. 

\subsection{Test Case Generation for Uninterpretable Model} 

In this subsection, we describe the various functions that are kept
undefined in Algorithm~\ref{algo:gse}.

\paragraph{Path Creation.} We start with the case of \texttt{getPath}
function. As illustrated in Section~\ref{sec:intro}, our key idea is
to use a local model generated from a local explainer -
LIME~\cite{LIME}. Our algorithm generates the decision tree instead of
the linear classifier as in LIME. 

\begin{algorithm}[h]
\scriptsize
Set$<$In,Out$>$ inout = localexpl(m,in)\\
return genDecionTree(inout);\\
\caption{getPath(Model m,Input in)}
\end{algorithm}

There are a few differences between the decision tree and a program
path.  1) Approximation: the decision tree path approximates the
actual execution path in terms of interpretable features. 2)
Confidence: there is no confidence associated with the predicates in a
program path, whereas a decision tree path has such confidence
associated with it. 3) Slice: decision tree may not contain some
predicates which do not have any effect to the resultant class. This
is analogous to dynamic slicing~\cite{DSLICE} in program analysis
literature. Such dynamic slicing removes the redundant state in
dynamic symbolic execution~\cite{B13}. 4) Non-repeatable: the local
explainer is not incremental in nature. In other words, it will not
try to preserve the predicates that have occurred in the same path. In
symbolic execution for software, predicates in the same path will
remain same. The effect of these differences is seen in our algorithm,
described subsequently.

\paragraph{Seed Input Selection.} 

Symbolic execution in program testing suffers from the path explosion
problem~\cite{G07}, especially the depth-first-search of explanation.
It can keep on exploring paths in the depth of the program tree,
without exploring the paths in the other parts of the program.
Researchers have explored various techniques to address this problem -
applying demand driven or directed technique which generates test
cases towards a particular location in the program, and compositional
techniques which try to analyze various functional modules separately
before combining them to generate longer paths in the whole program.
\eat{All these techniques exploit the structure of the program under
  test.}

To resolve the path explosion problem, we exploit the distribution
present in the training data (Function \texttt{seed\_test\_input()}).
Each training data instance can be a good starting point of the
symbolic search.  However, the order of training data instances become
important in search when there is a limit to the search due to which
executing all training instances is not possible.  Therefore, to
increase the diversity in search, we cluster the training data and
take seed inputs in round-robin fashion from each cluster. 

\begin{algorithm}[h]
\scriptsize
i = 0\\
clusters=KMeans(data,NumberOfClusters)\\
\While {i $<$ max(size(clusters))}{
 \If {q.qsize() == limit} {
     break
  }
  \ForEach {cluster $\in$ clusters}{
      \If {i $\ge$ get\_cluster\_size(cluster)}{
          continue\\
       }         
       row = cluster.rows.next\\
       rows.add(row)
  }
}
\Return(rows)
\caption{check\_for\_error\_condition()}
\end{algorithm}

\paragraph{Ranking.} Automated test case generation procedure is limit
based. It is therefore important to generate non-redundant and
effective test cases. We use a ranking scheme based on the confidence of
predicates in the decision tree to select which test input to execute
next. The confidence of the path is determined by the average
confidence of the predicates in it. Therefore, when we toggle a
predicate $p$, we consider the average confidence of the predicates in
the prefix of the path leading to and including $p$. This ranking
scheme orders the test inputs generated through the undirected symbolic
execution. In the next section, we discuss another place in the
algorithm for generating input (other than seed inputs and undirected
symbolic) and present relative ranking among them. 

\section{Checking Individual Discrimination}

Here we discuss few more changes to the generic algorithms above: 1)
checking the error condition, 2) directed search, and 3) relative
ranking.

\paragraph{Checking Individual Discrimination.}  We begin with the
case of checking individual discrimination.  Such check will occur in
function \texttt{check\_for\_error\_condition}.  The pseudo code for
this function is shown below. The algorithms make the check as per
definition of individual discrimination i.e. if a combination of
protected attribute values (from the domain) would result in any
different class for the non-protected attribute values contained in
the test case.

\begin{algorithm}[h]
\scriptsize
 class = model.test(t)\\
\ForEach {$\langle val_0,\ldots,val_n\rangle|val_i\in protected\_attribute_i.vals$}{
 tnew=Replace value of $protected\_attribute_i$ in $t$ with $val_i$ \\
 class1 = model.test(tnew)\\
 \If {class1 != class}{ 
    \Return Individual Discrimination Found
  }
}
\Return Individual Discrimination Not Found
\caption{check\_for\_error\_condition(t)}
\end{algorithm}

\begin{algorithm}[htb]
\scriptsize
\SetAlgoLined
\SetAlgoLined\DontPrintSemicolon
count = 0;\\\label{line:r1}
inputs = seed\_test\_inputs()\\\label{line:r2}
priorityQ q = empty; q.addAll(inputs,Rank1); \\\label{line:r3}
\While {count$<$limit \&\& !q.isEmpty()} { \label{line:r4}
  $t$ = deque(q)\\ \label{line:r5}
  found = check\_for\_error\_condition($t$) \\ \label{line:r5}
  Path p = getPath($t$)\\ \label{line:r6}
  \If {found}{
      // directed search\\
      \ForEach {predicate $c$ in order from top to bottom of path}{ \label{line:r8}
        path\_constraint = p.constraint  
        {\bf If} {$c$ is of protected attribute} {\bf then} continue\\
        \If {$c.confidence < T2$}{
           path\_constraint.remove(c)
  	   path\_constraint.add(not(c))\\ \label{line:r10}
	   \If {!visited\_path.contains(path\_constraint)}{
	       visited\_path.add(path\_constraint);\\
	       input= solve(path\_constraint)\\ \label{line:r11}
               r = average(path\_constraints)ï \\  \label{line:r12}
               q.add(input,Rank2-r)\\ \label{line:r13}
	    }
        }    	
     }
  } \label{line:r15}
  // undirected search\\
  prefix\_pred = true\\ \label{line:r16}
  \ForEach {predicate $c$ in order from top to bottom of path}{ \label{line:rl9}
        {\bf If} {$c$ is of protected attribute} {\bf then} continue\\
        {\bf If} {$c.confidence < T1$} {\bf then} break\\        
  	path\_constraint = prefix\_pred  $\wedge$ not (c)\\ \label{line:r20}
	\If {!visited\_path.contains(path\_constraint)}{
	   visited\_path.add(path\_constraint);\\
	   input= solve(path\_constraint)\\ \label{line:r21}
           r = average\_conf(path\_constraints)ï \\  \label{line:r21}
           q.add(input,Rank3+r)\\ \label{line:r11}
	}    
	prefix\_pred = prefix\_pred $\wedge$ c	\label{line:r22}
  }   
  count++ \label{line:r21}
}

\caption{Individual Discrimination}\label{algo:indi}
\end{algorithm}

\paragraph{Directed Symbolic Search.} The undirected symbolic search
(discussed previously) tries to find test inputs which can cause
individual discrimination. Once such a test case (say $t$) is found,
we try to generate more test inputs which can lead to the individual
discrimination. The key idea is to negate the low confidence
non-protected attribute constraint of $t$'s decision tree to generate
more test inputs. Low confidence constraints are less prone to change
the behavior of the test case and therefore can have the same effect
on protected attributes as in $t$.

The entire algorithm for individual discrimination is presented in
Algorithm~\ref{algo:indi}. Lines~\ref{line:r6}-\ref{line:r15} describe
the directed search whereas Lines~\ref{line:r16}-\ref{line:r22}
describe the undirected search. There are two major differences
between the directed and undirected search. The first difference is
that in a directed search only low confidence constraints are selected
for toggling because of the reason described above (Line 12).
In contrast, in undirected search, the high confidence constraints are
chosen for toggling. This is because high confidence constraint
toggling will result in more chance of diverse coverage of paths. The
second difference is that, in undirected search, no constraint exists
for suffix of the path, whereas in directed search, all constraints,
except the selected low confidence one to toggle, remain as it is.

\paragraph{Relative Ranking.} In the consolidated algorithm three
reference ranks are presented for seed input, directed search, and
undirected search. They are chosen in such a way that, the highest priority
is given to directed search followed by seed input and finally the
undirected search based on their ability to uncover discrimination
causing inputs.

In the next section, we experimentally show the effectiveness of
various optimizations described in this section.

\section{Experimental Evaluation}
\label{sec:expt}
\subsection{Setup}

\paragraph{Benchmark Characteristics.} We have used 8 open source
fairness benchmarks from various sources (see Table~\ref{tab:bench}).

\begin{table}[t]
\scriptsize
\begin{tabular}{|c|c|c|}
\hline
Benchmark & Size & Source \\
\hline
German Credit Data & 1000 & UCI Machine Learning Repository\\
Adult census income & 32561 &  UCI Machine Learning Repository\\
Bank marketing & 45211 & UCI Machine Learning Repository\\
US Executions & 1437 &  data.world - US Executions since 1977\\
Fraud Detection & 1100 & Kaggle - Fraud Detection\\
Raw Car Rentals & 486 & yelp.com - raw-car-rentals\\
credit data & 600 & modified  German Credit used in THEMIS\\
census data & 15360 & modified  Adult income used in THEMIS \\
\hline
\end{tabular}
\caption{Benchmark Characteristics\label{tab:bench}}
\end{table}

\paragraph{Configurations.} Our code is written in Python and executed
in Python 2.7.12 compiler. All experiments are performed in a machine
running Ubuntu 16.04, having 16GB RAM, 2.4Ghz CPU running Intel Core
i5. We have used LIME~\cite{LIME} for local explainability. We have
used KMeans clustering in training data with cluster size = 4. For
each benchmark, we have created Logistic regression with the default
configuration in scikit-learn. The selection of the model is inspired
by THEMIS. The two thresholds used in our algorithm T1=0.3 and T2=0.2
as shown in Algorithm~\ref{algo:indi}. 


\subsection{Experiments}

\paragraph{Goals.} Our experiments have two goals, given below:
\begin{itemize}
\item{Comparison with the existing work.} How well we perform compare
  to existing work in finding individual discrimination in models? We
  compare our system with the existing system called
  THEMIS~\cite{THEMIS} which checks the individual discrimination by
  random test case generation.
\item{Effect of Algorithmic Features.} How well each algorithmic
  feature (Directed and Undirected symbolic execution, training data)
  contributed to finding individual discrimination?
\end{itemize}

\begin{table}[h]
\scriptsize
\centering
\begin{tabular}{|l|l||r|r||r|r||}\hline
Bench. & Prot. Attr. &  \multicolumn{2}{|c||}{THEMIS} &  \multicolumn{2}{|c||}{Symbolic}\\\cline{3-6}
              &          & \#Gen & \#InDi & \#Gen & \#InDi\\\hline
German Credit &  gender & 999 & 166 & 1000 & 598 \\ 
\hline
German Credit &  age & 999 & 90 & 1000 & 359\\ 
\hline
Adult income &  race & 999 & 70 & 1000 &175\\ 
\hline
Adult income &  sex & 990 & 1 & 1000 & 462\\ 
\hline
Fraud Detection &  age & 999 & 3 & 656 & 0\\ 
\hline
Car Rentals &  Gender & 680 & 198 & 1000 & 801\\ 
\hline
credit &  i/gender & 598 & 44 & 1000 & 420 \\ 
\hline
census &  h/race & 999 & 57 & 1000 & 609\\ 
\hline
census &  i/sex & 999 & 7 & 1000 & 176\\ 
\hline
Bank Marketing &  age & 999 & 0 & 1000 & 1\\ 
\hline
US Executions &  Race & 999 & 2 & 1000 & 6\\ 
\hline
US Executions &  Sex & 999 & 8 &1000 & 31\\ 
\hline
\end{tabular}
\caption{Comparison with THEMIS}
\label{tab:themis}
\end{table}

\paragraph{Comparison to THEMIS.} To Compare with THEMIS, we got the
code from their GitHub repository and analyzed.  It seemed that there
is an unintended behavior. THEMIS actually produces duplicate test
cases and the result that is reported contains duplicate test cases.
We changed THEMIS's code to remove duplicates.  Table~\ref{tab:themis}
shows the result of the comparison to THEMIS.  $Gen$ refers to the set of
unique test cases generated. For each such test case, more test cases
are generated and executed to check the discrimination by changing the
value of protected attributes.  $InDi$ denotes the subset of the
generated test cases ($Gen$) which results in individual
discrimination. Note that, except in one case our algorithm produces
better results than THEMIS.

THEMIS has an average success score (\#Indi/\#Gen) of 6.4\% whereas
our symbolic algorithm has 30.3\% average success score.  It is
evident that \textbf{across 12 benchmarks, our algorithm generates 3.72
  times more} successful (that resulted in discrimination) test
cases than THEMIS.  This demonstrates advancement in the published
state-of-the-art in individual discrimination.

In Table~\ref{tab:themis2}, we report the contribution of the test
case generation feature (training data, undirected symbolic, directed
symbolic) contributed to the above success.  The result evidently
shows the effect of our relative ranking strategy which specifies the  decreasing order of preferences as Indirect, Data, and Direct. Note that,
on average, the success percentage for Data and Indirect Symbolic
execution are 23\% and 37\%, respectively.

\begin{table}
\centering
\scriptsize
\begin{tabular}{|l||r|r||r|r||r|r||}\hline
 Bench.  & \multicolumn{2}{|c|}{Data}  &  \multicolumn{2}{|c|}{Directed Symb.} & \multicolumn{2}{|c|}{UnDirected} \\\cline{2-7}
        &  Gen & InDi & Gen & InDi & Gen & Indi\\\hline
German Credit(gender) & 2 & 1 & 998 & 597 & 0 & 0 \\ 
\hline
German Credit(age) & 27 & 2 & 973 & 357 & 0 & 0 \\ 
\hline
Adult Income(race) & 313 & 24 & 687 & 151 & 0 & 0 \\ 
\hline
Adult Income(sex) & 77 & 12 & 923 & 450 & 0 & 0 \\ 
\hline
Fraud Detection & 1000 & 0 & 0 & 0 & 0 & 0 \\ 
\hline
Car Rentals & 18 & 15 & 982 & 786 & 0 & 0 \\ 
\hline
credit & 20 & 1 & 980 & 419 & 0 & 0 \\ 
\hline
census(race) & 1 & 1 & 999 & 608 & 0 & 0 \\ 
\hline
census(sex) & 41 & 2 & 959 & 174 & 0 & 0 \\ 
\hline
Bank Marketing & 984 & 1 & 16 & 0 & 0 & 0 \\ 
\hline
US Executions(Race) & 877 & 0 & 22 & 4 & 101 & 2 \\ 
\hline
US Executions(Sex) & 877 & 0 & 74 & 29 & 49 & 2 \\ 
\hline
\end{tabular}
\caption{Contribution of different features}
\label{tab:themis2}
\end{table}


\paragraph{Importance of Training Data.} We conducted two experiments
for determining the importance of training data by changing the seed input
function which instead of taking training data, takes random data from the domain. In the first experiment (shown in Table~\ref{tab:data1}) we
switch off the directed and undirected symbolic execution so that all
the test cases are generated from the seed data. In this second
experiment (Table~\ref{tab:data2}, we keep both the symbolic searches.
In the first experiment, we see that just by getting that training
data we get an average improvement of 108\% (25\% to 12\%). The second
experiment shows the effectiveness of symbolic evaluation even if we
start with the random input. In the \texttt{credit} data
(Table~\ref{tab:data2}), we see that random data got 310 successful
test cases and is less effective than training data which has got 421
successful test cases. However, it's much better than THEMIS (44).

\begin{table}
\centering
\scriptsize
\begin{tabular}{|l|l||r|r||r|r||}\hline
 Bench.  & \multicolumn{2}{|c|}{Training Data} & \multicolumn{2}{|c|}{Random}\\\hline
        &  Gen & InDi & Gen & InDi\\\hline
 credit  &  500 & 56 & 500 & 25 \\ \hline
 German (Age) & 1000 & 70 & 1000 & 46 \\ \hline
Census (Sex) & 500 & 20 & 500 & 5\\ \hline
Car & 500 & 394 & 500 & 190\\ \hline
\end{tabular}
\caption{Training data as seed (w/o Symbolic)}
\label{tab:data1}
\end{table}

\begin{table}
\centering
\scriptsize
\begin{tabular}{||l||r|r|r|r||}\hline
 Bench.  &   \multicolumn{4}{|c||}{Random} \\\hline
        &  Total & Seed & Dir. & UNdirect. \\ \hline
credit  & 310/1000 & 1/21 & 309/979 & 0/0 \\ \hline
German (Age) & 365/100 & 4/49 & 361/951 & 0/0\\ \hline
Census (Sex) & 195/1000 & 3/87 & 192/913 & 0/0 \\ \hline
Car & 803/1000 & 14/21 & 789/979 & 0/0\\\hline
\end{tabular}
\caption{Random seed data (with symbolic) (\#InDi/\#Gen)}
\label{tab:data2}
\end{table}

\paragraph{Importance of Directed Search} Based on the previous
experiments we notice that directed symbolic search has high
percentage (37\%) of effectiveness. We conduct another experiment to
see how we directed search affects the overall execution of the
algorithm given a limit on the number of test cases (1000). The
results by switching off the directed search feature is presented in
Table~\ref{tab:dir}. We should compare this result with the results in
Table~\ref{tab:themis2} for the 4 benchmarks. We see that the average
effectiveness drops from 43.9\% to 25.2\% for these 4 benchmarks by
removing the directed search feature. This shows the importance of the
directed search technique. 

\begin{table}
\centering
\scriptsize
\begin{tabular}{|l|r|r|r|r||}\hline
 Bench.  &   \multicolumn{4}{|c|}{Random} \\\hline
        &  Total & Seed & Dir. & Un-direct. \\ \hline
credit  & 66/603 & 66/600 & 0/0 & 0/3 \\ \hline
German (Age) & 70/1000 & 70/1000 & 0/0 & 0/0\\ \hline
Census (Sex) & 45/994 & 45/992 & 0/0 & 0/2\\ \hline
Car & 231/295 & 91/114 & 0/0 & 140/181\\ \hline
\end{tabular}
\caption{Without directed search (\#InDi/\#Gen)}
\label{tab:dir}
\end{table}

\begin{figure}[t]
\centering
\includegraphics[height=1.5in]{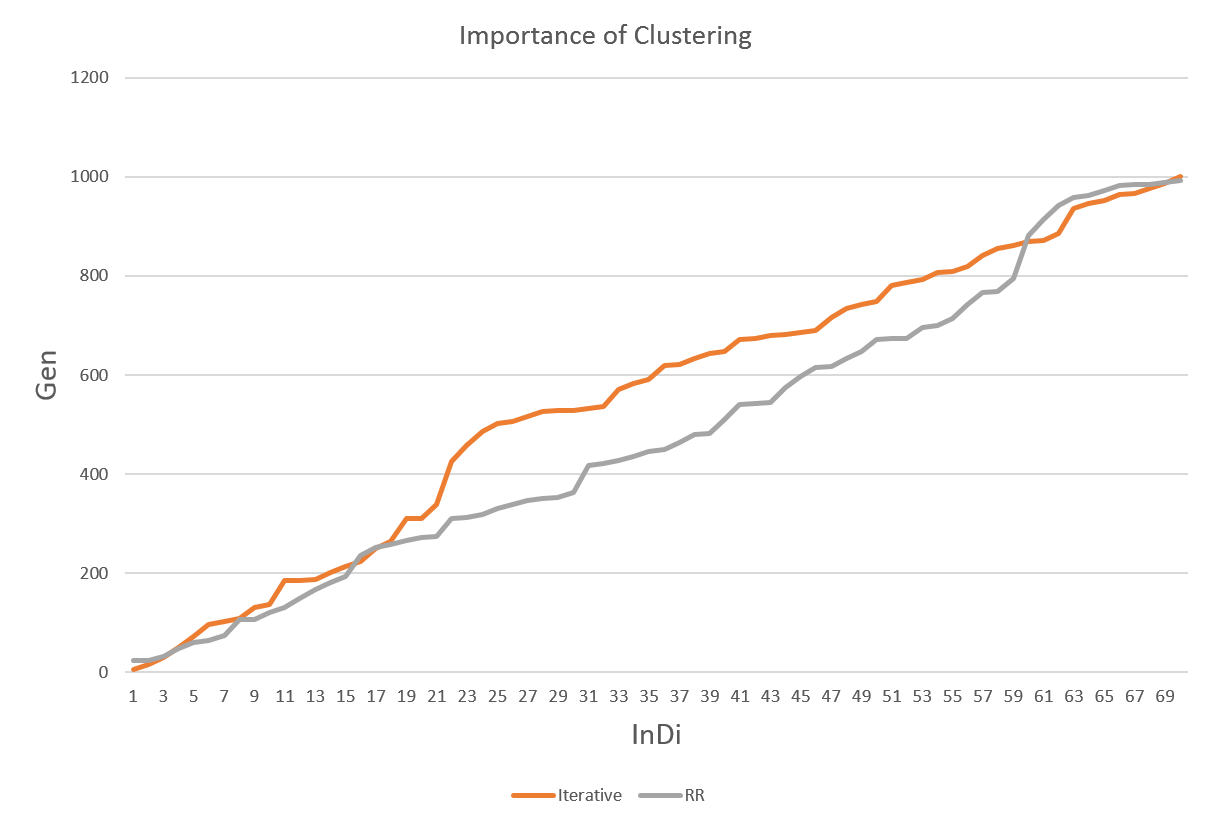}
\caption{Importance of clustering (German-age)}
\label{fig:clus}
\end{figure}

\paragraph{Importance of Clustering} Figure~\ref{fig:clus} shows the
use of diverse seed data ordering got getting individual
discrimination. When the number of test cases is limited then we expect
that diverse ordering (round-robin) will fetch more discrimination
than iterative ordering. It is evident from the Figure~\ref{fig:clus}
that, for most number of test cases, the number of discrimination
found is higher for round-robin selection from clusters is more
effective than iterative selection. For example, in the execution of 600
test cases, the round-robin found 45 discrimination compared to 35 for
iterative.

\paragraph{Importance of Undirected Search} We performed an experiment
by giving highest priority for undirected search and removing directed
search feature. We see that in two cases (German-age and Car) the
undirected search extracted successful test cases (see
Table~\ref{tab:undir}). In the other two cases, very fewer test cases are
even generated 4 and 7 as the confidence of predicates is not high for
toggling.

\begin{table}
\centering
\scriptsize
\begin{tabular}{|l|r|r|r|r||}\hline
 Bench.  &   \multicolumn{4}{|c|}{Random} \\\hline
        &  Total & Seed & Dir. & Un-direct. \\ \hline
credit  & 66/607 & 66/600 & 0/0 & 0/7 \\ \hline
German (Age) & 18/1000 & 0/1 & 0/0 & 18/999\\ \hline
Census (Sex) & 45/1000 & 45/996 & 0/0 & 0/4\\ \hline
Car & 254/325 & 75/97 & 0/0 & 179/228\\ \hline
\end{tabular}
\caption{Undirected search (w/o directed)  (\#InDi/\#Gen)}
\label{tab:undir}
\end{table}

Overall, our experiments demonstrate that directed search uncovers
many bias instances after finding the fault. For initial fault finding, training data works better than the undirected symbolic search.

\section{Related Work}
\label{sec:related}

We present the related works in two categories - testing of models and
detecting individual discrimination.

\paragraph{Automated Test Case Generation of AI Models.} 

We discuss the works which perform symbolic/concolic based test case
generation of AI models. $DeepCheck$~\cite{DEEPCHECK} uses a white box
technique which performs symbolic execution on deep neural networks
with the target of generating adversarial images. $DeepCheck^{\tau}$
translates the network $\mathcal{N}$ to an imperative program
$\mathcal{P}$ that has the same behavior as the neural network
$\mathcal{N}$. $DeepCheck^{Imp}$ executes the program $\mathcal{P}$
and for the execution path $I$, taken in $\mathcal{P}$, finds the
important pixels by 1) first finding a linear expression in terms of
the input variables and 2) assigns scores to the input pixels based on
the coefficient in the expression, 3) selects the important pixels
from the top threshold. A new image is created by changing the $t$
important pixel such that the label changes. The problem with the
technique is that the semantic preserving translation mechanism only
works for a specific network. Concolic execution~\cite{DART} (Concrete
and Symbolic) on deep neural networks is performed by
\cite{DEEPCONCOLIC}. Their technique is white box and goal is to
perform coverage of deep neural network by systematic test case
generation. They model the network using linear constraints and use a
specialized solver to generate test cases.  Wicker et
al.~\cite{FEATURE} aim to cover the input space by exhaustive mutation
testing that has theoretical guarantees, while
in~\cite{DEEPXPLORE,DEEPTEST,DEEPGAUGE} gradient-based search
algorithms are applied to solve optimization problems, and Sun et
al.~\cite{TESTINGDNN} apply linear programming.

All of the above techniques are white box compare to our black box
technique. We use an off-the-shelf solver to generate test cases.
Compare to other approaches which try to consider test generation for
creating adversarial input in the image space, our technique addresses
a new problem trust and ethics domain.

\paragraph{Individual Discrimination} 

THEMIS~\cite{THEMIS} uses the causality to define individual
discrimination. Even though they use a black box technique, their test
case generation technique uses random test generation instead of any
systematic test case generation. In fact, they envision the use of
systematic test case generation techniques in their paper.
FairTest~\cite{FAIRTEST} uses manually written tests to measure four
kinds of discrimination scores. Their idea is to use indirect
co-relation between attributes (e.g., salary is related to age) to
generate test cases. FairML~\cite{FAIRML} uses an iterative procedure,
based on an orthogonal projection of input attributes, for enabling
interpretability of black-box predictive models. Through an iterative
procedure, one can quantify the relative dependence of a black-box
model on its input attributes. The relative significance of the inputs
to a predictive model can then be used to assess the fairness (or
discriminatory extent) of such a model.

None of the existing individual discrimination technique uses
systematic test case generation, even though all such methods are
black box techniques. Our is the first method which uses systematic
test case generation for individual discrimination with the advantage
of black box method.

\section{Conclusion}
\label{sec:conc}

In this paper, we present a test case generation algorithm for checking
individual discrimination in AI models. Our approach combines the idea
of symbolic evaluation which systematically generates test inputs and
local explanation which approximates the path in the model using
linear models. The resultant technique is black box. In future, we
would like to apply this technique in various domains including text
and images. We would also like to measure the efficacy of symbolic
execution in models using the structural metric like neuron coverage,
boundary value coverage, etc.

\newpage
\bibliography{bib}
\bibliographystyle{aaai}

\end{document}